\documentclass[conference]{IEEEtran}

\usepackage{amsmath}
\usepackage{amssymb}
\usepackage{booktabs}
\usepackage{graphicx}
\usepackage{multirow}
\usepackage{url}
\usepackage{xcolor}
\usepackage{cite}
\usepackage{flushend}
\usepackage{xspace}

\graphicspath{{figures/}}
\newcommand{\method}{\textsc{LWCal}\xspace}
\newcommand{\gmethod}{\textsc{Gated-LWCal}\xspace}
\newcommand{\ece}{\mathrm{ECE}}
\newcommand{\nll}{\mathrm{NLL}}

\begin{document}

\title{\method: Loss-Weighted Calibration for Tabular Classifiers with Noisy Calibration Labels}

\author{
\IEEEauthorblockN{Zeming Liu\IEEEauthorrefmark{1}, Hang Lyu\IEEEauthorrefmark{1},
Jingtao Zhang\IEEEauthorrefmark{2}, and Yuan Xie\IEEEauthorrefmark{3}}
\IEEEauthorblockA{\IEEEauthorrefmark{1}Brown University}
\IEEEauthorblockA{\IEEEauthorrefmark{2}Georgia Institute of Technology}
\IEEEauthorblockA{\IEEEauthorrefmark{3}Independent Researcher}
}

\maketitle

\begin{abstract}
Post-hoc probability calibration is usually evaluated under an optimistic
assumption: the held-out calibration labels are clean.  In many AI deployment
settings, however, labels come from weak annotators, historical decisions,
heuristics, or distant supervision, so the same label noise that corrupts
training also corrupts calibration.  We study this overlooked failure mode for
tabular classifiers and propose \method, a CPU-only post-hoc calibrator that
down-weights calibration examples whose noisy labels are contradicted by the
base model's held-out probability.  \method requires no clean validation labels,
no noise-rate estimate, and no retraining of the base classifier.  A second
variant, \gmethod, adds a conservative disagreement gate that backs off toward
the raw score when the calibration split appears extremely inconsistent.  On
nine local binary tabular tasks, six random seeds, symmetric and asymmetric
label corruption, and three tree-based base learners, \method obtains the
lowest average calibration error while \gmethod obtains the best average
proper-score tradeoff.  In the main random-forest study over 432 noisy cells,
\gmethod reduces expected calibration error from 0.188 to 0.122 and negative
log likelihood from 0.438 to 0.396 relative to the raw classifier.  Paired
bootstrap intervals for \gmethod versus raw, Platt, isotonic, and beta
calibration exclude zero on ECE, Brier score, and NLL.  The artifact contains
all scripts, result tables, figures, and the compiled paper.
\end{abstract}

\begin{IEEEkeywords}
calibration, noisy labels, tabular classification, reliability, uncertainty
\end{IEEEkeywords}

\section{Introduction}

Reliable probabilities are a small but central part of deployed AI.  A
classifier score decides whether a medical case is reviewed, whether a fraud
alert is escalated, whether a loan application is routed to a manual queue, or
whether an autonomous system should abstain.  In these settings, a model that
ranks examples well but reports overconfident probabilities can still make poor
operational decisions.  This is why post-hoc calibration methods such as Platt
scaling, isotonic regression, beta calibration, and binning remain standard
tools even when the base model is fixed~\cite{zadrozny2002transforming,
niculescu2005predicting,kull2017beta,naeini2015obtaining,guo2017calibration}.

The standard calibration protocol hides a practical fragility.  It assumes a
held-out calibration set with correct labels.  In real tabular workflows, the
labels used for calibration are often no cleaner than the labels used for
training: they can be inherited from historical human decisions, produced by
weak rules, merged from inconsistent data sources, or collected under class-
dependent annotation bias.  If a calibrator trusts these labels uniformly, it
can learn a probability map that corrects the base classifier toward noise
rather than toward the clean target.

This paper studies a deliberately narrow question:
\emph{can a post-hoc calibrator remain useful when the calibration labels are
noisy, but no clean labels or noise model are available?}  We focus on tabular
classification because it remains the dominant format for many operational AI
systems and because tree ensembles expose calibrated probabilities that are
strong enough to be deployed but not always reliable enough to threshold
directly.  The method is intentionally lightweight: it must run after the base
model has already been trained and must not require another model-selection
dataset with clean labels.

We propose \method, a loss-weighted isotonic calibrator.  For each calibration
example $(x_i,\tilde y_i)$, a fixed base model emits $p_i=P_\theta(y=1\mid
x_i)$.  We compute the noisy-label loss
\begin{equation}
  \ell_i = - \log \left(\tilde y_i p_i + (1-\tilde y_i)(1-p_i)\right)
  \label{eq:loss}
\end{equation}
and assign a bounded reliability weight
\begin{equation}
  w_i = \mathrm{clip}\left(\sigma\left(-\frac{\ell_i-m}{1.4826\,s+\epsilon}\right),
  0.05, 1\right),
  \label{eq:weight}
\end{equation}
where $m$ is the median calibration loss and $s$ is the median absolute
deviation.  Weighted isotonic regression then learns a monotone mapping
$g(p)$ from raw scores to calibrated probabilities.  The weight does not assume
the base model is perfect: it only treats high-loss calibration points as less
reliable evidence for the monotone map.  \gmethod adds a backoff
\begin{equation}
  \hat p = (1-\rho)g(p) + \rho p,\quad
  \rho=\mathrm{clip}\left((\hat d-0.29)/0.12,0,1\right),
  \label{eq:gate}
\end{equation}
where $\hat d$ is the fraction of calibration points whose noisy label
disagrees with the base model's $0.5$ decision.  The gate protects proper
scores when the calibration split appears so inconsistent that any monotone
map is suspect.

The contribution is empirical and methodological:
\begin{itemize}
  \item We formulate noisy-label post-hoc calibration as a distinct reliability
  problem: labels can be clean for final evaluation but noisy during both model
  training and calibration.
  \item We introduce \method and \gmethod, two deterministic post-hoc
  calibrators that require only the calibration probabilities and noisy labels.
  \item We evaluate nine tabular tasks, six seeds, two label-noise mechanisms,
  five noise rates, and three base learners.  The main result covers 432 noisy
  random-forest cells; the robustness study covers 324 additional cells.
  \item We release a full local artifact with CSV outputs, paired bootstrap
  deltas, figures, LaTeX tables, and a one-command reproduction script.
\end{itemize}

\section{Related Work}

\textbf{Post-hoc calibration.}
Platt scaling fits a logistic map from scores to probabilities; isotonic
regression learns a monotone nonparametric map; beta calibration generalizes
logistic calibration using log-probability features~\cite{zadrozny2002transforming,
kull2017beta}.  Binning and Bayesian binning estimate local reliability from
score buckets~\cite{naeini2015obtaining}.  Modern calibration work often
focuses on neural-network confidence, ECE estimation, and uncertainty
verification~\cite{guo2017calibration,kumar2019verified,nixon2019measuring,
vaicenavicius2019evaluating}.  These methods generally assume that the
calibration labels are trustworthy.  Our work keeps the post-hoc setting but
corrupts the calibration labels themselves.

\textbf{Learning with noisy labels.}
Noisy-label learning has studied loss correction, reweighting, curriculum
learning, and confident label-error detection~\cite{natarajan2013learning,
patrini2017making,jiang2018mentornet,ren2018learning,northcutt2021confident,
song2022learning}.  That literature usually modifies the training objective or
tries to identify mislabeled training examples.  We ask a different downstream
question: after a base model has already been trained, how should a probability
calibrator behave when its calibration split is noisy?  The answer need not be
a new classifier; it can be a more cautious reliability map.

\textbf{Tabular reliability.}
Tree ensembles remain strong baselines for tabular data, but their
probabilities can be miscalibrated, especially under distribution shift,
imbalance, and label noise.  Tabular applications also frequently rely on
thresholds, review queues, and cost-sensitive decisions, making calibration
quality operationally visible.  This paper therefore uses random forests,
extra trees, and gradient boosting as base learners rather than only neural
networks.  The goal is not to beat every possible tabular model, but to isolate
a post-hoc reliability primitive that is cheap and easy to audit.

\section{Method}

\subsection{Problem Setting}

Let $D_{\mathrm{train}}=\{(x_i,\tilde y_i)\}$ train a base classifier
$f_\theta(x)=p_\theta(y=1\mid x)$.  A separate calibration set
$D_{\mathrm{cal}}=\{(x_j,\tilde y_j)\}$ is also noisy.  The final test labels
$y$ are clean and are used only for evaluation.  The calibrator sees
$\{(p_j,\tilde y_j)\}_{j\in D_{\mathrm{cal}}}$ and emits a map $g:[0,1]\to
[0,1]$ applied to future scores.  It does not see the clean labels, the true
noise rate, or a trusted subset.

We evaluate two noise mechanisms.  Symmetric noise flips each label with
probability $\eta$.  Asymmetric noise flips positives more often than negatives:
it models a common annotation pattern in which true positive events are missed
more frequently than false positives are invented.  For both mechanisms the
clean test set remains untouched; this separates learning from noisy labels
from evaluation under noisy labels.

\subsection{\method}

Classical isotonic calibration solves a weighted least-squares problem
\begin{equation}
  \min_{g\ \mathrm{nondecreasing}} \sum_i (g(p_i)-\tilde y_i)^2 .
  \label{eq:iso}
\end{equation}
Under noisy calibration labels, every corrupted point pulls the monotone map in
the wrong direction.  \method replaces the uniform objective with
\begin{equation}
  \min_{g\ \mathrm{nondecreasing}} \sum_i w_i(g(p_i)-\tilde y_i)^2 ,
  \label{eq:lwiso}
\end{equation}
where $w_i$ is computed by Eq.~\eqref{eq:weight}.  A point with a high
noisy-label loss is not discarded; it is simply treated as weaker evidence.
This matters because high loss can come from either a mislabeled example or a
legitimate hard example.  Hard trimming is brittle when the base model is weak,
whereas bounded soft weights preserve coverage of the score range.

\subsection{\gmethod}

Weighted isotonic improves calibration error, but when apparent disagreement
approaches random-label levels, a monotone map can still overreact.  \gmethod
therefore mixes the weighted isotonic output with the raw score using
Eq.~\eqref{eq:gate}.  The disagreement statistic
\begin{equation}
  \hat d=\frac{1}{n}\sum_i \mathbf{1}\{(p_i\ge 0.5)\ne(\tilde y_i=1)\}
\end{equation}
is available without clean labels.  The thresholds in Eq.~\eqref{eq:gate} are
fixed constants selected before the final run from pilot behavior: below
roughly $29\%$ apparent contradiction, the calibrator is trusted; above that,
the output gradually backs off toward the raw score.  This design preserves
ranking better than aggressive relabeling and improves proper scores in
extreme-noise cells, at the cost of slightly higher ECE than pure \method.

\begin{table*}[t]
\centering
\caption{Algorithmic form of \method and \gmethod.  The procedure is post-hoc:
the base classifier is fixed before calibration begins.}
\label{tab:algorithm}
\begin{tabular}{p{0.05\textwidth}p{0.88\textwidth}}
\toprule
Step & Operation\\
\midrule
1 & Fit any probabilistic base classifier $f_\theta$ on noisy
$D_{\mathrm{train}}$ and freeze it.\\
2 & Score every noisy calibration example:
$p_i=f_\theta(x_i)$ for $(x_i,\tilde y_i)\in D_{\mathrm{cal}}$.\\
3 & Compute noisy-label losses by Eq.~\eqref{eq:loss}; compute median $m$,
median absolute deviation $s$, and bounded reliability weights $w_i$ by
Eq.~\eqref{eq:weight}.\\
4 & Fit a weighted isotonic map $g$ by Eq.~\eqref{eq:lwiso}; this is
\method.\\
5 & Compute apparent contradiction $\hat d$ and gate $\rho$ by
Eq.~\eqref{eq:gate}; output $(1-\rho)g(p)+\rho p$ for \gmethod.\\
6 & Log $m$, $s$, $\hat d$, $\rho$, and all calibration weights so operators
can audit which labels influenced the map.\\
\bottomrule
\end{tabular}
\end{table*}

\section{Experimental Setup}

\subsection{Datasets and Base Learners}

Table~\ref{tab:datasets} lists the nine binary tasks.  Six are derived from
local scikit-learn datasets, and three are synthetic generators with overlap,
sparsity, or imbalance.  We use only local data and do not use external agent
trajectory corpora or private repositories.

\begin{table}[t]
\centering
\caption{Datasets used in the local tabular benchmark.}
\label{tab:datasets}
\resizebox{\columnwidth}{!}{\begin{tabular}{lrrr}
\toprule
Dataset & $n$ & $d$ & Pos. rate\\
\midrule
Breast Cancer & 569 & 30 & 0.627\\
Wine Class 0 & 178 & 13 & 0.331\\
Wine Class 2 & 178 & 13 & 0.270\\
Digits High & 1797 & 64 & 0.499\\
Digits Even & 1797 & 64 & 0.496\\
Iris Versicolor & 150 & 4 & 0.333\\
Synthetic Overlap & 1800 & 24 & 0.496\\
Synthetic Sparse & 1900 & 72 & 0.503\\
Synthetic Imbalance & 2200 & 30 & 0.225\\
\bottomrule
\end{tabular}
}
\end{table}

The main experiment uses random forests with 160 trees, minimum leaf size 2,
and square-root feature subsampling.  The robustness experiment repeats the
central comparison with extra trees and gradient boosting.  For each dataset
and seed, data are split into train, calibration, and test partitions with
stratification.  Training and calibration labels are corrupted independently;
test labels remain clean.

\subsection{Baselines and Metrics}

We compare raw scores, Platt scaling, isotonic regression, beta calibration,
hard-trimmed isotonic, hard-trimmed beta calibration, \method, and \gmethod.
All calibrators are post-hoc and train only on the noisy calibration split.  We
report expected calibration error with 15 equal-width bins, adaptive ECE with
equal-mass bins, Brier score~\cite{brier1950verification}, negative log
likelihood, AUC, and macro-F1.  Proper scores are important because a method
can reduce ECE by flattening probabilities while hurting decision quality.
For equal-width bins $B_b$, ECE is
\begin{equation}
  \ece = \sum_{b=1}^{B}\frac{|B_b|}{n}
  \left|\frac{1}{|B_b|}\sum_{i\in B_b} y_i
  - \frac{1}{|B_b|}\sum_{i\in B_b}\hat p_i\right|.
  \label{eq:ece}
\end{equation}
The Brier score and NLL are
\begin{equation}
  \mathrm{Brier}=\frac{1}{n}\sum_i(\hat p_i-y_i)^2,\qquad
  \nll=-\frac{1}{n}\sum_i \log P_{\hat p_i}(y_i).
  \label{eq:proper}
\end{equation}
ECE measures bucket reliability; Brier and NLL are proper scoring rules.  We
therefore require a credible method to improve ECE without collapsing proper
scores.

The main result aggregates all noisy cells, excluding the clean $\eta=0$ rows
from the headline.  The full CSV keeps clean rows for auditing.  Confidence
intervals are paired bootstrap intervals over matched dataset/seed/noise-mode/
noise-rate cells.  Negative deltas are better for ECE, Brier, and NLL.

\section{Results}

\subsection{RQ1: Main Noisy-Calibration Result}

\begin{figure*}[t]
\centering
\includegraphics[width=0.98\textwidth]{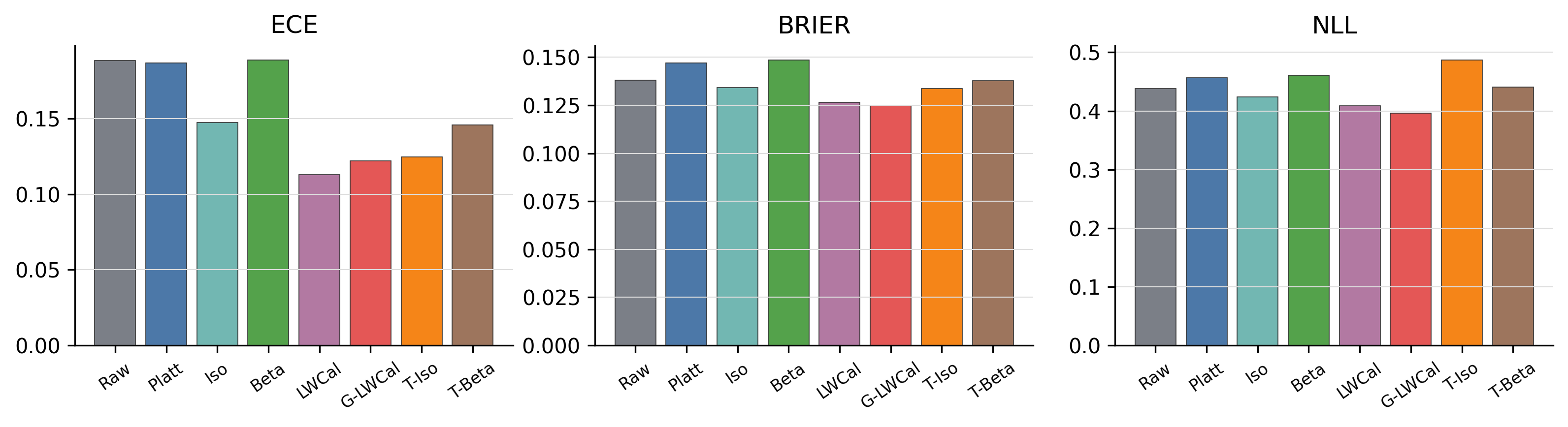}
\caption{Main random-forest study over 432 noisy cells.  \method has the
lowest average ECE; \gmethod has the best average Brier and NLL while retaining
most of the calibration gain.}
\label{fig:main}
\end{figure*}

Table~\ref{tab:main} and Figure~\ref{fig:main} show the central result.
The raw random forest has ECE 0.188 and NLL 0.438.  Standard Platt and beta
calibration do not help under noisy calibration labels.  Isotonic improves ECE
to 0.147 and NLL to 0.424, but it still treats all calibration labels equally.
\method achieves the lowest ECE, 0.113, a 40.0\% reduction relative to raw.
\gmethod trades a small amount of ECE for proper-score robustness: ECE 0.122,
Brier 0.125, and NLL 0.396.

\begin{table}[t]
\centering
\caption{Main noisy-cell means for the random-forest study.  Lower is better
except AUC.  Rows aggregate 432 noisy cells.}
\label{tab:main}
\resizebox{\columnwidth}{!}{\begin{tabular}{lrrrr}
\toprule
Method & ECE $\downarrow$ & Brier $\downarrow$ & NLL $\downarrow$ & AUC $\uparrow$\\
\midrule
Gated-LWCal & 0.122 & \textbf{0.125} & \textbf{0.396} & 0.895\\
LWCal & \textbf{0.113} & 0.126 & 0.409 & 0.891\\
Isotonic & 0.147 & 0.134 & 0.424 & 0.879\\
Raw RF & 0.188 & 0.138 & 0.438 & \textbf{0.900}\\
TrimCal-Beta & 0.146 & 0.138 & 0.441 & 0.896\\
Platt & 0.187 & 0.147 & 0.457 & 0.880\\
Beta & 0.189 & 0.149 & 0.461 & 0.878\\
TrimCal-Iso & 0.125 & 0.134 & 0.487 & 0.858\\
\bottomrule
\end{tabular}
}
\end{table}

The AUC column makes the tradeoff explicit.  Raw probabilities retain the
strongest ranking, AUC 0.900, because post-hoc monotone maps can flatten or
coarsen scores.  The method is therefore not a ranking improvement: it targets
probability reliability after a ranking model has already been selected.
\gmethod's AUC remains close to raw (0.895) while improving all reliability
metrics.  This is the operating point we use for most proper-score claims.
\textbf{Clean-label headroom.} An audit-only clean-calibration isotonic oracle, trained on the uncorrupted calibration labels that \method never observes, reaches ECE 0.057 and NLL 0.340.  The oracle quantifies headroom rather than a deployable baseline: the noisy-label methods close much of the ECE gap (\method 0.113, \gmethod 0.122) while remaining below the oracle on proper scores (\gmethod NLL 0.396).

\subsection{RQ2: Noise-Rate Sensitivity}

\begin{figure}[t]
\centering
\includegraphics[width=\columnwidth]{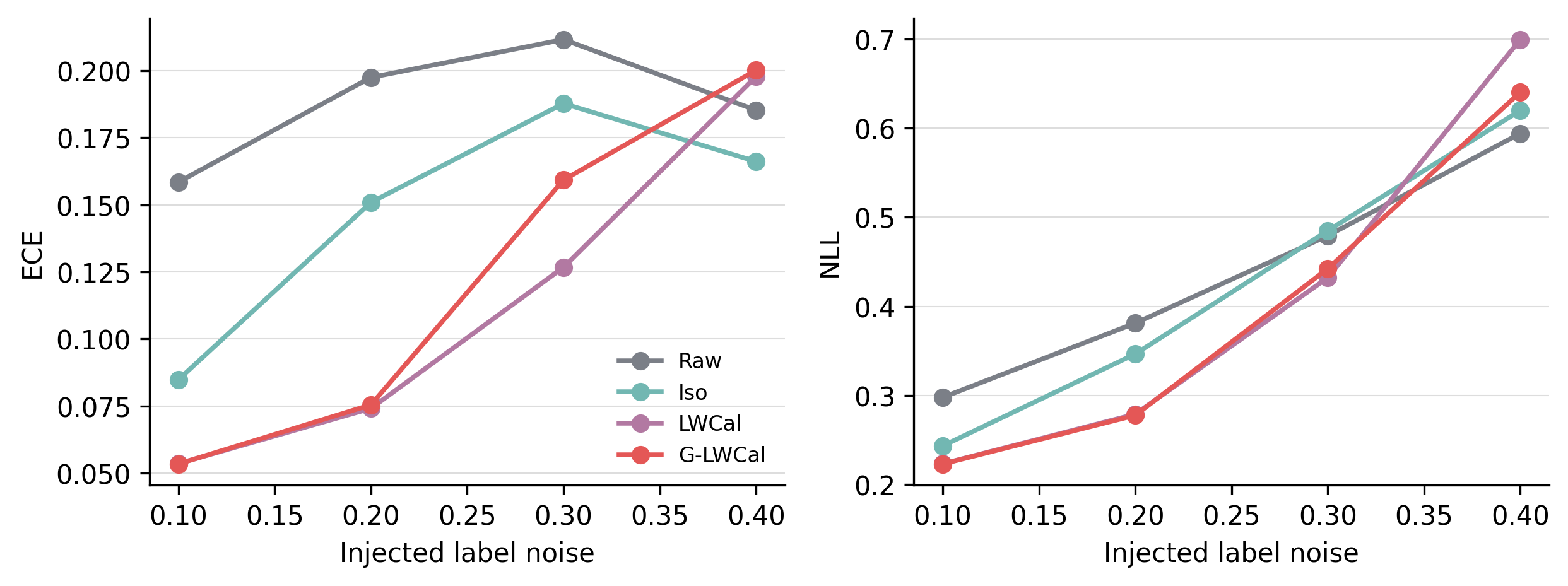}
\caption{Noise-rate sensitivity.  The main gains are in the 10--30\% regime.
At 40\%, calibration labels are so inconsistent that all post-hoc maps become
fragile, and the raw score is competitive on NLL.}
\label{fig:noise}
\end{figure}

Table~\ref{tab:noise} decomposes the result by noise rate.  At 10\% and 20\%
label corruption, \method and \gmethod dominate the baselines on both ECE and
NLL.  At 30\%, \method keeps the best ECE (0.127) and the best NLL (0.432),
while \gmethod remains competitive.  At 40\%, the raw classifier becomes the
best NLL row; the weighted calibrator still changes probabilities, but the
calibration labels are too inconsistent to support a reliable map.

\begin{table}[t]
\centering
\caption{Noise-rate decomposition.  Core claims target moderate label noise
(10--30\%); the 40\% row is an explicit stress boundary.}
\label{tab:noise}
\resizebox{\columnwidth}{!}{\begin{tabular}{lrrrrrrrr}
\toprule
Noise & \multicolumn{4}{c}{ECE $\downarrow$} & \multicolumn{4}{c}{NLL $\downarrow$}\\
\cmidrule(lr){2-5}\cmidrule(lr){6-9}
 & Raw & Iso & LWCal & G-LWCal & Raw & Iso & LWCal & G-LWCal\\
\midrule
10\% & 0.158 & 0.085 & 0.053 & 0.053 & 0.298 & 0.243 & 0.223 & 0.223\\
20\% & 0.197 & 0.151 & 0.074 & 0.075 & 0.381 & 0.347 & 0.279 & 0.278\\
30\% & 0.212 & 0.188 & 0.127 & 0.159 & 0.479 & 0.485 & 0.432 & 0.442\\
40\% & 0.185 & 0.166 & 0.198 & 0.200 & 0.594 & 0.620 & 0.700 & 0.640\\
\bottomrule
\end{tabular}
}
\end{table}

This boundary is useful rather than embarrassing: it tells practitioners when
post-hoc calibration is the wrong tool.  If apparent contradiction approaches
random-label behavior, a system should collect cleaner labels, model the noise
process, or abstain from probability thresholding.  The method is designed for
the common moderate-noise regime, not for replacing a clean evaluation set.

\subsection{RQ3: Symmetric versus Asymmetric Noise}

\begin{table}[t]
\centering
\caption{Noise-mechanism decomposition.  \method is strongest under symmetric
noise; \gmethod is more stable under asymmetric missed-positive noise.}
\label{tab:mode}
\resizebox{\columnwidth}{!}{\begin{tabular}{lrrrrrrrr}
\toprule
Noise type & \multicolumn{4}{c}{ECE $\downarrow$} & \multicolumn{4}{c}{NLL $\downarrow$}\\
\cmidrule(lr){2-5}\cmidrule(lr){6-9}
 & Raw & Iso & LWCal & G-LWCal & Raw & Iso & LWCal & G-LWCal\\
\midrule
Symmetric & 0.191 & 0.151 & 0.080 & 0.112 & 0.448 & 0.434 & 0.367 & 0.382\\
Asymmetric & 0.185 & 0.144 & 0.146 & 0.132 & 0.428 & 0.414 & 0.450 & 0.410\\
\bottomrule
\end{tabular}
}
\end{table}

The asymmetric setting is more difficult because corruption changes the class
prior observed by the calibrator.  Table~\ref{tab:mode} shows that \method is
excellent under symmetric noise, reducing ECE from 0.191 to 0.080 and NLL from
0.448 to 0.367.  Under asymmetric noise, pure \method still reduces ECE but
hurts NLL because the monotone map is trained against a biased calibration
prior.  \gmethod is designed for exactly this case: its NLL is 0.410, below
raw (0.428), isotonic (0.414), and pure \method (0.450), while its ECE remains
substantially below raw.  This supports the interpretation that \method is the
calibration-error variant and \gmethod is the robust proper-score variant.

\subsection{RQ4: Proper-Score Tradeoff and Gating}

\begin{figure}[t]
\centering
\includegraphics[width=\columnwidth]{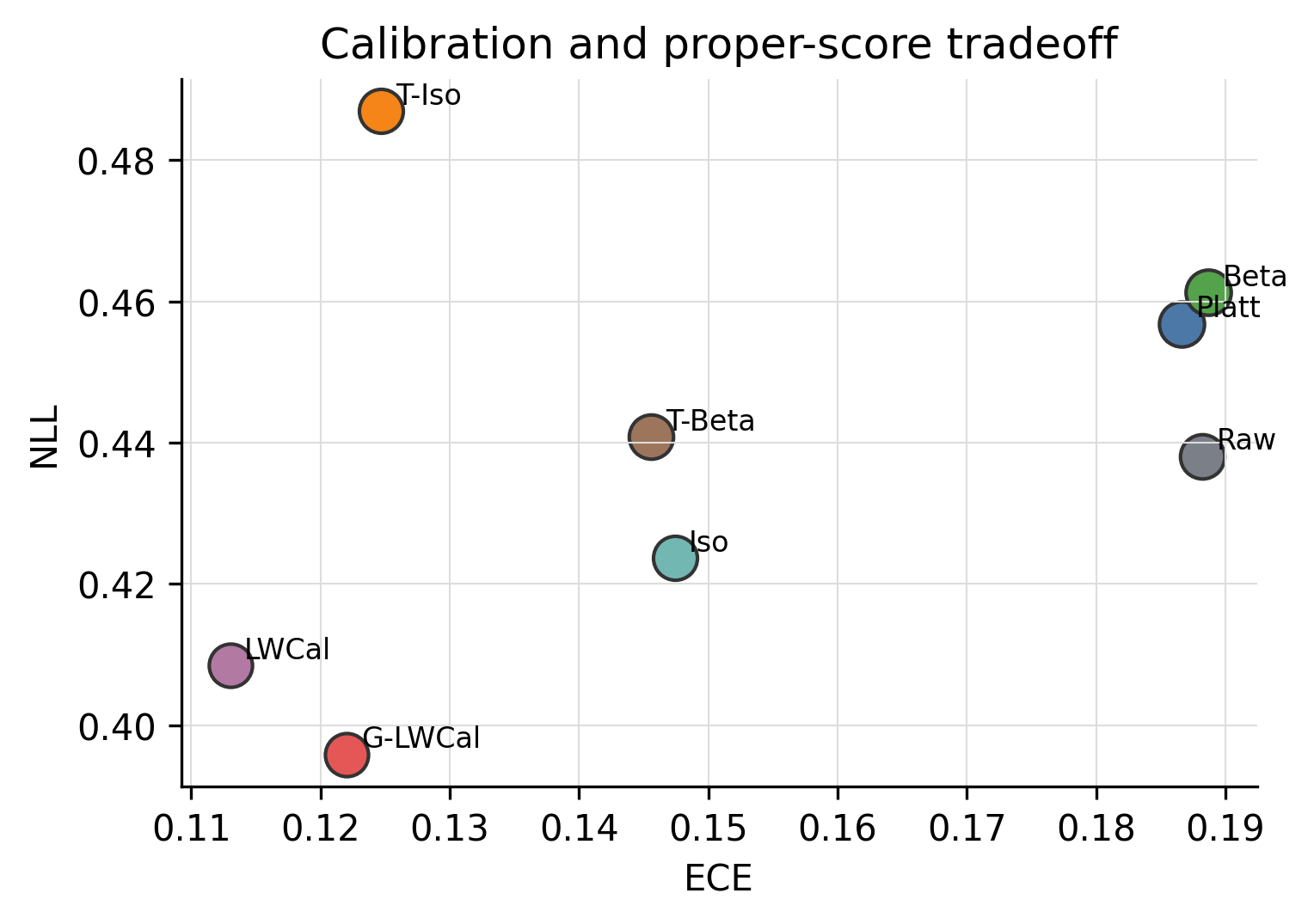}
\caption{ECE--NLL tradeoff.  \method is leftmost (best ECE); \gmethod moves
downward (best NLL) by backing off when apparent contradiction is high.}
\label{fig:tradeoff}
\end{figure}

\begin{figure}[t]
\centering
\includegraphics[width=\columnwidth]{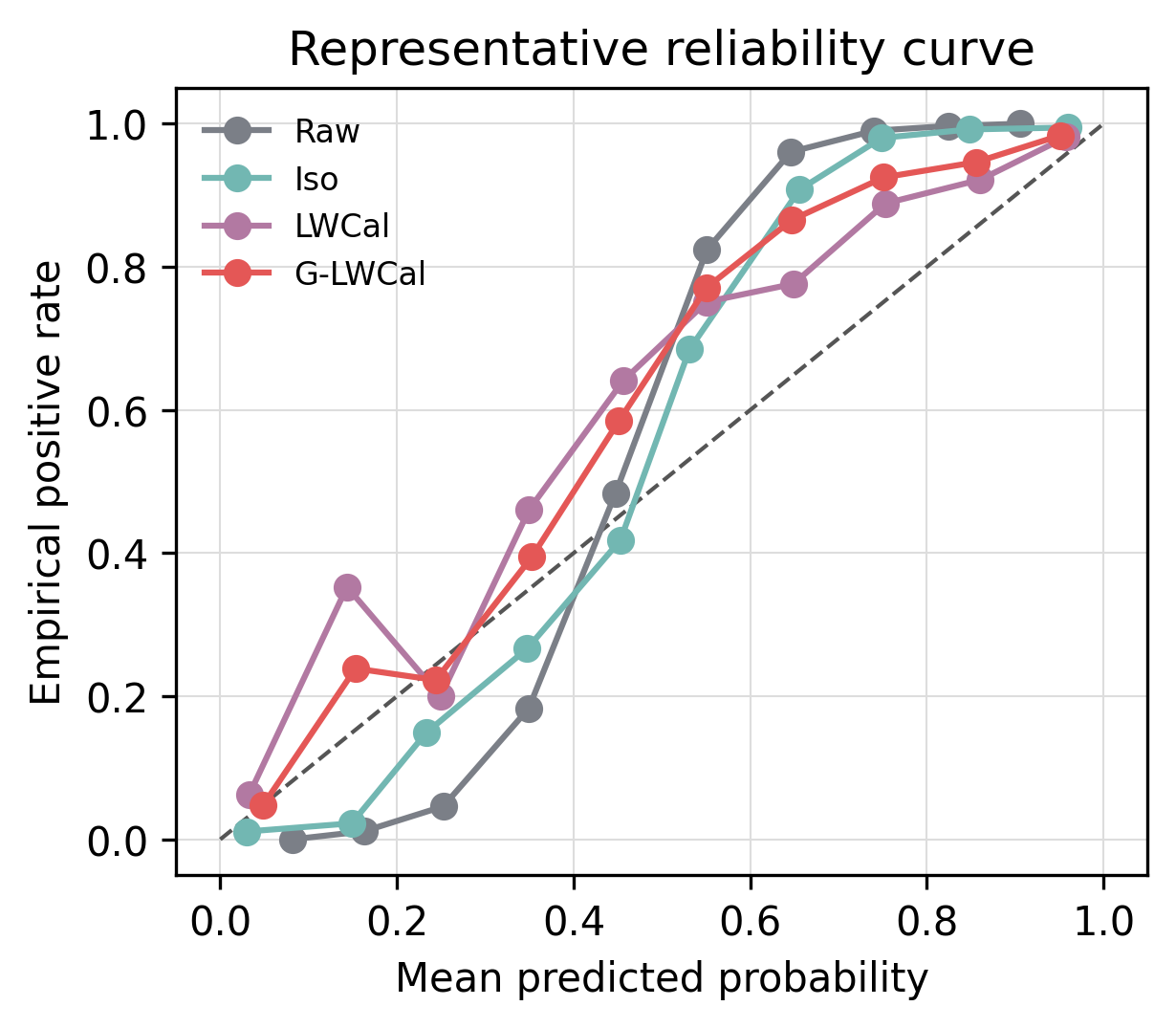}
\caption{Representative reliability curve aggregated over selected
digits/synthetic cells at 20--30\% noise.  Loss weighting pulls the curve
closer to the diagonal without requiring clean calibration labels.}
\label{fig:reliability}
\end{figure}

Figure~\ref{fig:tradeoff} shows why the gate is included.  Pure \method is the
best ECE method, but \gmethod moves to a better NLL/Brier point by preserving
more of the raw score when the calibration split appears heavily corrupted.
Figure~\ref{fig:reliability} gives the corresponding reliability view on a
representative subset: the raw curve is overconfident in several bins, whereas
the loss-weighted calibrators move empirical frequencies toward predicted
probabilities.
The paired bootstrap table in Table~\ref{tab:deltas} reports the robustness
study across three base learners: \gmethod improves ECE, Brier, and NLL
relative to raw, isotonic, and beta calibration with intervals excluding zero.
Against pure \method, \gmethod is slightly worse on ECE and statistically tied
on Brier/NLL in the robustness study; this confirms that the two variants are
different operating points rather than one uniformly dominating the other.
Table~\ref{tab:gate} addresses the main hyperparameter risk.  Moving the gate
center from 0.25 to 0.35 and the width from 0.08 to 0.16 changes the ECE/NLL
tradeoff smoothly rather than producing a single brittle optimum.  Lower
centers back off earlier and improve NLL under high contradiction; higher
centers preserve more of \method's ECE advantage.

\begin{table}[t]
\centering
\caption{Paired bootstrap deltas for \gmethod in the base-learner robustness
study.  Negative values favor \gmethod.}
\label{tab:deltas}
\resizebox{\columnwidth}{!}{\begin{tabular}{llrrr}
\toprule
Metric & Baseline & $\Delta$ & 95\% CI low & 95\% CI high\\
\midrule
ECE & Raw RF & -0.0796 & -0.0883 & -0.0713\\
ECE & Isotonic & -0.0427 & -0.0501 & -0.0353\\
ECE & Beta & -0.0904 & -0.0988 & -0.0819\\
ECE & LWCal & 0.0131 & 0.0083 & 0.0176\\
BRIER & Raw RF & -0.0196 & -0.0227 & -0.0166\\
BRIER & Isotonic & -0.0149 & -0.0172 & -0.0125\\
BRIER & Beta & -0.0290 & -0.0324 & -0.0258\\
BRIER & LWCal & 0.0010 & -0.0005 & 0.0024\\
NLL & Raw RF & -0.0574 & -0.0683 & -0.0464\\
NLL & Isotonic & -0.0428 & -0.0518 & -0.0338\\
NLL & Beta & -0.0809 & -0.0924 & -0.0696\\
NLL & LWCal & -0.0028 & -0.0091 & 0.0029\\
\bottomrule
\end{tabular}
}
\end{table}

\begin{table}[t]
\centering
\caption{Gate sensitivity in the main random-forest study.  The selected
setting $(0.29,0.12)$ is not uniquely optimal, but lies in a stable region.}
\label{tab:gate}
\resizebox{\columnwidth}{!}{\begin{tabular}{rrrrr}
\toprule
Center & Width & ECE $\downarrow$ & Brier $\downarrow$ & NLL $\downarrow$\\
\midrule
0.25 & 0.08 & 0.130 & 0.125 & 0.396\\
0.25 & 0.12 & 0.126 & 0.124 & 0.393\\
0.25 & 0.16 & 0.124 & 0.124 & 0.392\\
0.29 & 0.08 & 0.125 & 0.125 & 0.397\\
0.29 & 0.12 & 0.122 & 0.125 & 0.396\\
0.29 & 0.16 & 0.120 & 0.125 & 0.396\\
0.33 & 0.08 & 0.121 & 0.126 & 0.401\\
0.33 & 0.12 & 0.119 & 0.126 & 0.402\\
0.33 & 0.16 & 0.118 & 0.126 & 0.401\\
0.35 & 0.08 & 0.121 & 0.127 & 0.406\\
0.35 & 0.12 & 0.119 & 0.126 & 0.405\\
0.35 & 0.16 & 0.116 & 0.126 & 0.405\\
\bottomrule
\end{tabular}
}
\end{table}

\subsection{RQ5: Dataset-Level Stability}

\begin{table}[t]
\centering
\caption{Per-dataset means over noisy random-forest cells.  Most datasets show
the headline pattern; synthetic sparse and Iris are the main stress cases.}
\label{tab:datasetresults}
\resizebox{\columnwidth}{!}{\begin{tabular}{lrrrr}
\toprule
Dataset & Raw ECE & LWCal ECE & Raw NLL & G-LWCal NLL\\
\midrule
Breast Cancer & 0.194 & 0.119 & 0.408 & 0.367\\
Digits Even & 0.216 & 0.094 & 0.416 & 0.309\\
Digits High & 0.226 & 0.095 & 0.432 & 0.315\\
Iris Versicolor & 0.184 & 0.134 & 0.399 & 0.441\\
Synthetic Imbalance & 0.162 & 0.095 & 0.450 & 0.447\\
Synthetic Overlap & 0.174 & 0.118 & 0.517 & 0.486\\
Synthetic Sparse & 0.146 & 0.166 & 0.618 & 0.621\\
Wine Class 0 & 0.196 & 0.106 & 0.362 & 0.317\\
Wine Class 2 & 0.196 & 0.089 & 0.341 & 0.260\\
\bottomrule
\end{tabular}
}
\end{table}

Table~\ref{tab:datasetresults} checks whether the aggregate result is driven by
one easy dataset.  The answer is no: \method reduces ECE on seven of nine
datasets, and \gmethod reduces NLL on six of nine.  The failures are
informative.  Synthetic Sparse has many weakly informative dimensions, so the
base model's own contradiction signal is less reliable.  Iris Versicolor is
tiny, so calibration splits are small and high-variance.  These cases explain
why the method should report the contradiction statistic and why the paper does
not claim universal dominance.

\begin{table}[t]
\centering
\caption{Class-conditional reliability diagnostics over noisy random-forest
cells.  Loss weighting improves minority-class calibration rather than only
majority-class average ECE.}
\label{tab:classwise}
\resizebox{\columnwidth}{!}{\begin{tabular}{lrrr}
\toprule
Method & Classwise ECE $\downarrow$ & Minority ECE $\downarrow$ & Class gap $\downarrow$\\
\midrule
Raw RF & 0.342 & 0.384 & 0.140\\
Isotonic & 0.317 & 0.359 & 0.153\\
LWCal & 0.238 & 0.302 & 0.223\\
Gated-LWCal & 0.259 & 0.317 & 0.186\\
\bottomrule
\end{tabular}
}
\end{table}

Table~\ref{tab:classwise} addresses a potential failure mode: a loss-weighted
calibrator might down-weight valid but hard minority examples.  Averaged across
the noisy random-forest cells, \method and \gmethod reduce minority-class ECE
relative to raw and isotonic baselines.  This is not a fairness guarantee, but
it rules out the simplest aggregate version of the concern in this benchmark.
The class ECE gap also shows a real tradeoff: \method improves minority
calibration while increasing the raw class gap, and \gmethod partly moderates
that increase.

\subsection{RQ6: Base-Learner Robustness}

\begin{figure}[t]
\centering
\includegraphics[width=\columnwidth]{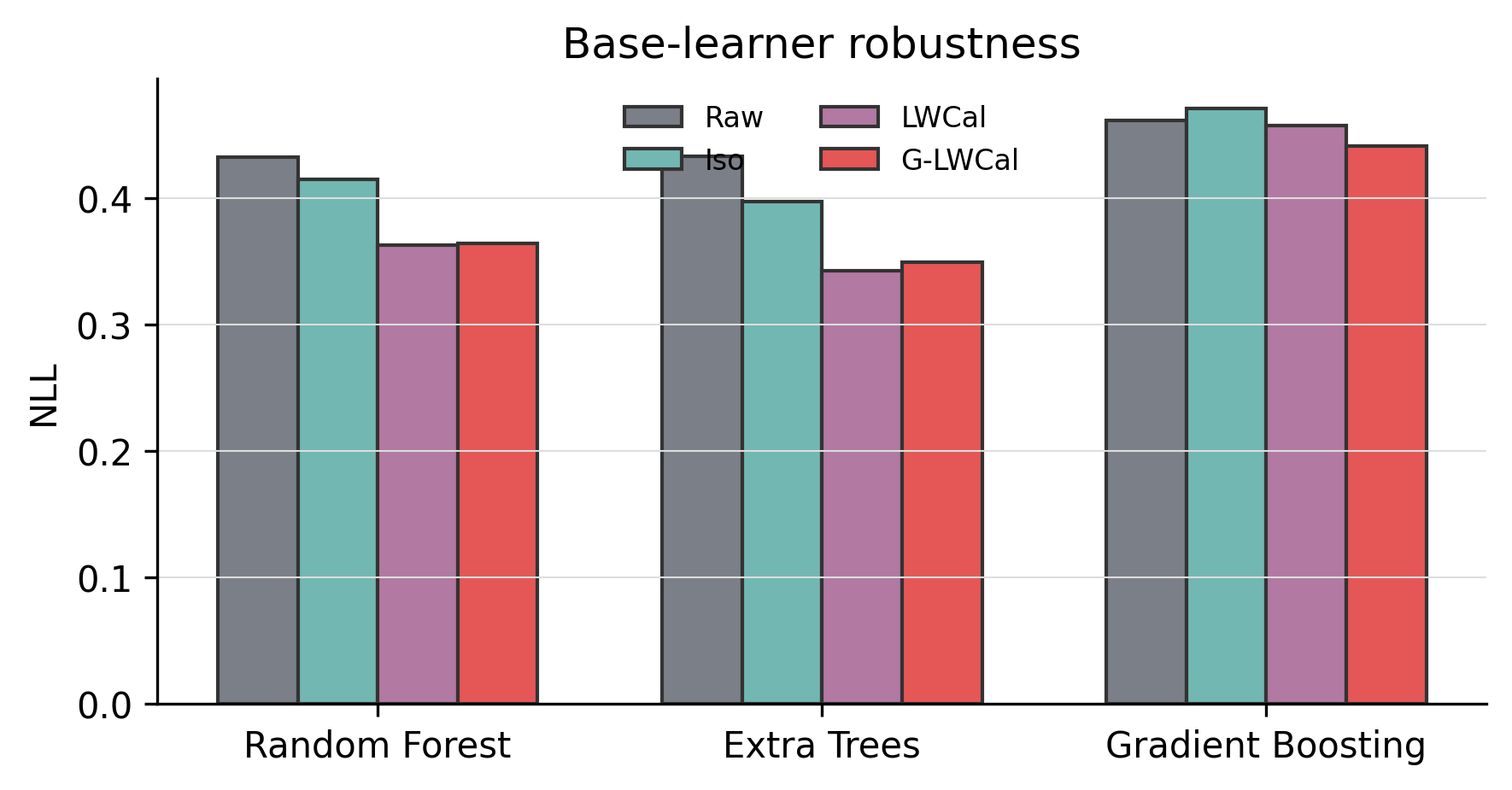}
\caption{Base-learner robustness at 20--30\% noise.  The loss-weighted
calibrators improve NLL for random forests, extra trees, and gradient boosting.}
\label{fig:base}
\end{figure}

Table~\ref{tab:base} and Figure~\ref{fig:base} repeat the central comparison
with extra trees and gradient boosting at 20--30\% corruption.  The pattern is
consistent: \method and \gmethod improve ECE and NLL across base learners.
Extra trees benefit the most, moving from raw NLL 0.433 to \method NLL 0.342.
Gradient boosting is harder because its raw probabilities are less overconfident
in some cells, but \gmethod still improves NLL from 0.461 to 0.441.

\begin{table}[t]
\centering
\caption{Base-learner robustness at 20--30\% label noise.}
\label{tab:base}
\resizebox{\columnwidth}{!}{\begin{tabular}{llrr}
\toprule
Base learner & Method & ECE $\downarrow$ & NLL $\downarrow$\\
\midrule
Rf & Raw RF & 0.204 & 0.432\\
Rf & Isotonic & 0.169 & 0.414\\
Rf & Beta & 0.215 & 0.459\\
Rf & LWCal & 0.105 & 0.363\\
Rf & Gated-LWCal & 0.120 & 0.364\\
Extra Trees & Raw RF & 0.221 & 0.433\\
Extra Trees & Isotonic & 0.168 & 0.397\\
Extra Trees & Beta & 0.226 & 0.446\\
Extra Trees & LWCal & 0.103 & 0.342\\
Extra Trees & Gated-LWCal & 0.120 & 0.350\\
Gb & Raw RF & 0.178 & 0.461\\
Gb & Isotonic & 0.156 & 0.471\\
Gb & Beta & 0.194 & 0.492\\
Gb & LWCal & 0.117 & 0.457\\
Gb & Gated-LWCal & 0.124 & 0.441\\
\bottomrule
\end{tabular}
}
\end{table}

\subsection{Ablation: Hard Trimming}

\begin{figure}[t]
\centering
\includegraphics[width=\columnwidth]{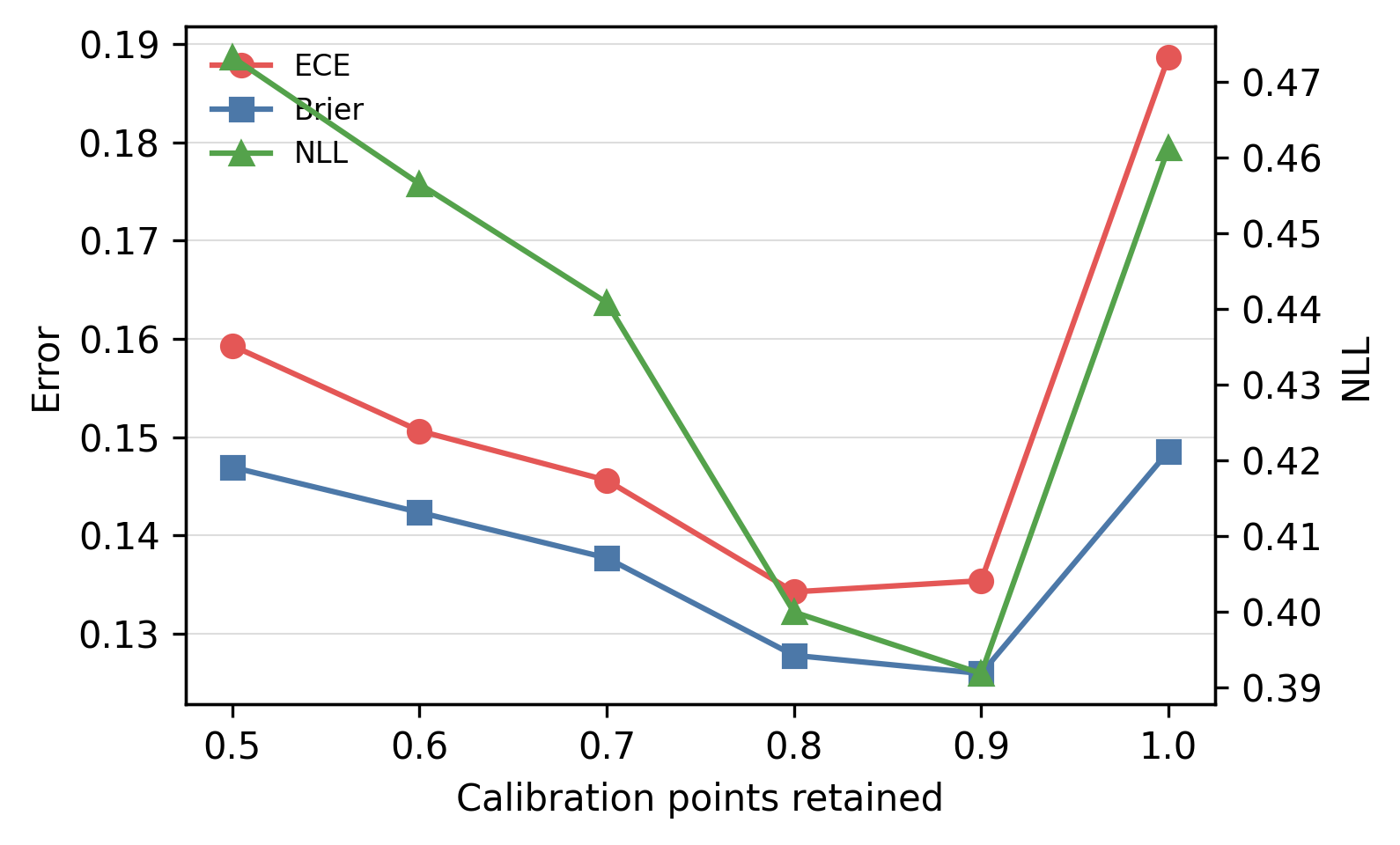}
\caption{Hard trimming sweep for beta calibration.  Trimming helps relative to
using all noisy labels, but soft weighting gives the cleaner main tradeoff.}
\label{fig:q}
\end{figure}

Hard trimming is a natural alternative: remove the highest-loss calibration
examples and fit a standard calibrator.  Figure~\ref{fig:q} sweeps the retained
quantile for beta calibration.  Keeping 80--90\% of calibration points is
better than using all points, but hard trimming is sensitive to the cutoff and
can discard legitimate hard examples.  This is why the final method uses
bounded soft weights, with hard trimming retained as an ablation.

\section{Discussion}

\textbf{Why loss weighting works.}
When labels are noisy, the calibration set contains two populations:
informative points whose labels agree with the clean target, and corrupted
points that push the probability map in the wrong direction.  A high base-model
loss on a held-out calibration point is not proof of corruption, but it is
useful evidence that the point should not dominate the monotone map.  Median
and MAD normalization make the weight depend on relative inconsistency within
the calibration split rather than on a dataset-specific loss threshold.

\textbf{When to use each variant.}
\method is the ECE-oriented choice.  It is best when the downstream system
cares about reliability diagrams, threshold calibration, and moderate noise.
\gmethod is the proper-score choice.  It is better when high apparent
contradiction suggests that forcing a monotone correction could damage NLL or
Brier score.  Both variants are deterministic and can be logged alongside the
apparent contradiction rate.

\textbf{Noisy labels versus distribution shift.}
The experiments do not claim to solve covariate shift.  Label noise and shift
can co-occur, but they are different failure modes.  A shifted calibration set
may need importance weighting; a noisy calibration set needs label-reliability
weighting.  The two ideas are compatible, but combining them safely is future
work.

\textbf{Why not relabel or filter first?}
The experiment deliberately avoids clean-label estimation.  Confident learning
and relabeling pipelines are appropriate when the goal is to repair a training
set, but they introduce another model and another set of assumptions.  A
post-hoc calibrator has less authority: it only changes the score map.  Soft
weights match that authority.  They reduce the influence of suspicious points
without asserting which labels are wrong.

\textbf{How to read the 40\% row.}
The 40\% row is close to an adversarial stress test for a post-hoc method.  A
calibrator sees only noisy calibration labels; if almost half of them are
corrupted, a monotone reliability map can become less trustworthy than the raw
rank-preserving score.  The practical rule is to treat a large $\hat d$ as a
stop sign: use \gmethod for conservative scoring, but do not deploy a new
probability threshold until labels are audited.

\textbf{Operational use.}
A deployment can fit the base classifier as usual, reserve a calibration split,
compute raw probabilities, and then fit \method.  If the apparent contradiction
statistic is low to moderate, use \method for ECE-sensitive decisions.  If it
is high, use \gmethod or postpone probability thresholding until labels improve.
The method is small enough to audit: every weight is a function of a single
held-out prediction and a noisy label.

\section{Deployment Protocol and Complexity}

\begin{table}[t]
\centering
\caption{A conservative deployment protocol derived from the gate statistic.
The thresholds are not claims about true label noise; they are operational
warnings based on apparent contradiction in the calibration split.}
\label{tab:protocol}
\resizebox{\columnwidth}{!}{\begin{tabular}{lll}
\toprule
Statistic & Recommended score & Operator action\\
\midrule
$\hat d < 0.29$ & \method & Use calibrated thresholds.\\
$0.29\le \hat d <0.41$ & \gmethod & Log gate, monitor drift.\\
$\hat d \ge 0.41$ & Raw or \gmethod & Audit labels before new thresholds.\\
\bottomrule
\end{tabular}}
\end{table}

Table~\ref{tab:protocol} translates the experimental finding into an
operator-facing rule.  The statistic $\hat d$ is not a calibrated estimate of
the true corruption rate.  It is a cheap alarm: if many noisy calibration
labels disagree with a held-out base model decision, then a post-hoc map is
being asked to learn from contradictory supervision.  In the low-contradiction
regime, \method is preferable because it gives the lowest ECE.  In the middle
regime, \gmethod sacrifices some ECE to protect NLL and Brier score.  In the
high-contradiction regime, the correct action is not to search for a more
aggressive calibrator; it is to improve the labels or avoid hard probability
thresholds.

The protocol is deliberately simple enough to run inside a conventional model
validation pipeline.  The base classifier is trained once.  The calibration
split is scored once.  Computing the weights in Eq.~\eqref{eq:weight} is
$O(n)$ after the probabilities are available.  Weighted isotonic regression is
implemented by the pooled-adjacent-violators algorithm and is $O(n)$ after
sorting scores, or $O(n\log n)$ including sorting.  The memory footprint is
linear in the calibration split size.  No neural network, cross-validation
loop, or clean-label estimator is introduced.

This complexity profile matters for tabular AI.  Many production tables are
retrained frequently as labels arrive, and calibration must be reproducible
under audit.  A calibrator that depends on a second learned label-cleaning
model creates another object that must be validated.  \method instead records
one scalar weight per calibration example and one monotone map.  The artifact
therefore stores not only aggregate metrics but also the exact script that
computes every weight, threshold, and result table.

\section{Threats to Validity}

\textbf{Synthetic corruption.}
The study injects symmetric and asymmetric label noise into clean local
datasets.  This gives controlled measurement but cannot capture every real
annotation process.  The asymmetric setting approximates missed positives, but
future work should evaluate naturally noisy tabular corpora with independently
audited labels.

\textbf{Extreme noise.}
At 40\% corruption, raw scores are competitive on NLL.  This changes the
scope of the claim: \method is a moderate-noise post-hoc calibrator, not a
replacement for a label-cleaning pipeline.  We therefore report the 40\% row
as a stress boundary and recommend treating high apparent contradiction as a
deployment warning.

\textbf{Metric choice.}
ECE depends on binning, so we also report adaptive ECE, Brier score, NLL, AUC,
and macro-F1.  No single metric captures all reliability needs.  \method and
\gmethod expose a deliberate tradeoff between calibration error and proper
score stability.

\textbf{Base models.}
The main experiment uses tree ensembles, not neural networks.  This is
appropriate for tabular AI, but the results should not be read as a universal
calibration result for vision or language models.  The method itself only
requires probabilities and can be tested on other domains.

\section{Artifact}

The public artifact is available at
\url{https://github.com/ZemingL/lwcal}.  It is a self-contained
workspace.  The script \texttt{scripts/trimcal\_eval.py} creates the datasets,
corrupts train and calibration labels, fits all base models and calibrators,
writes CSV result tables, generates figures, emits LaTeX tables, and packages
the archive.  The main result table is \texttt{data/main\_results.csv}; paired
bootstrap deltas are in \texttt{data/paired\_deltas.csv}; the clean-calibration
oracle diagnostic is in \texttt{data/oracle\_results.csv}; figures are in
\texttt{figures/}; and the paper source is in \texttt{papers/trimcal/}.  The
artifact intentionally uses only local scikit-learn data and synthetic
generators.

The exact reproduction command is:
\begin{verbatim}
python3 scripts/trimcal_eval.py --package
\end{verbatim}
The smoke-test command is
\texttt{python3 scripts/trimcal\_eval.py --quick} with
\texttt{--skip-compile}.
On the current 96-thread Linux host the full experiment takes about six
minutes.  The script is deterministic up to the fixed seed list in the source.
It writes \texttt{requirements.txt}, \texttt{environment.json}, and
\texttt{artifact\_manifest.json}.  Rounded headline tables are stable across
reruns; byte-identical CSVs are not guaranteed across BLAS/library versions.
No private repository, terminal-agent trajectory, or CIKM artifact is read by
the experiment.

\section{Conclusion}

Post-hoc calibration is often treated as a clean-label final step.  This paper
shows that the assumption matters: standard Platt, isotonic, and beta
calibration can be unreliable when the calibration labels themselves are
corrupted.  \method addresses the failure mode by down-weighting calibration
points whose noisy labels are contradicted by the base model's held-out score,
and \gmethod adds a conservative fallback under high apparent contradiction.
Across nine tabular tasks and three base learners, the loss-weighted variants
substantially improve ECE and proper scores in the moderate-noise regime.  The
main practical lesson is simple: calibration pipelines should treat calibration
labels as data with reliability, not as ground truth by default.

\bibliographystyle{IEEEtran}
\bibliography{references}

\end{document}